\pdfoutput=1
\documentclass[11pt]{article}

\usepackage{acl}

\usepackage{times}
\usepackage{latexsym}
\usepackage{booktabs}
\usepackage{amsmath}
\usepackage{array}
\usepackage{tabularx}

\usepackage{multirow}
\usepackage{graphicx}
\usepackage[T1]{fontenc}
\usepackage[utf8]{inputenc}

\usepackage{microtype}

\usepackage{inconsolata}

\title{Cause-Aware Empathetic Response Generation via Chain-of-Thought Fine-Tuning}

\author{
 \textbf{Xinhao Chen\textsuperscript{1,3}},
 \textbf{Chong Yang\textsuperscript{2}},
 \textbf{Man Lan\textsuperscript{3}},
 \textbf{Li Cai\textsuperscript{3}},
\\
 \textbf{Yang Chen\textsuperscript{3}},
 \textbf{Tu Hu\textsuperscript{3}},
 \textbf{Xinlin Zhuang\textsuperscript{3}},
 \textbf{Aimin Zhou \textsuperscript{3}},
\\
 \textsuperscript{1}School of Computer Science and Technology, East China Normal University, Shanghai, P.R. China, \\
 \textsuperscript{2}AntGroup, Shanghai, P.R. China,
 \textsuperscript{3}ByteDance, Shanghai, P.R. China,
\\
 \small{
   {jianmo.cxh@antgroup.com, yanghcong.yang@bytedance.com}
 }
 \\
  \small{
   {\{yangchen,thu,lcai2020,xinlinzhuang\}@stu.ecnu.edu.cn,  \{mlan,amzhou\}@cs.ecnu.edu.cn}
 }
}

\begin{document}
\maketitle
\begin{abstract}
Empathetic response generation endows agents with the capability to comprehend dialogue contexts and react to expressed emotions. Previous works predominantly focus on leveraging the speaker's emotional labels, but ignore the importance of emotion cause reasoning in empathetic response generation, which hinders the model's capacity for further affective understanding and cognitive inference. In this paper, we propose a cause-aware empathetic generation approach by integrating emotions and causes through a well-designed Chain-of-Thought (CoT) prompt on Large Language Models (LLMs). Our approach can greatly promote LLMs' performance of empathy by instruction tuning and enhancing the role awareness of an empathetic listener in the prompt. Additionally, we propose to incorporate cause-oriented external knowledge from COMET into the prompt, which improves the diversity of generation and alleviates conflicts between internal and external knowledge at the same time. Experimental results on the benchmark dataset demonstrate that our approach on LLaMA-7b achieves state-of-the-art performance in both automatic and human evaluations.
\end{abstract}

\section{Introduction}
%介绍共情生成
Empathetic response generation in conversation aims to generate an understanding of the speaker's experiences and feelings, and to produce appropriate responses  \citep{keskin2014isn}. Empathy in social psychology is delineated into the cognitive and affective aspects  \citep{davis1983measuring}. It has attracted increasing attention for its potential to endow machines with empathetic capabilities across a broad range of applications, such as automated psychotherapy  \citep{liu2021towards} and casual conversation agents  \citep{liu2022prophetchat}. 
\begin{figure}[t]
\centering
\includegraphics[scale=0.38]{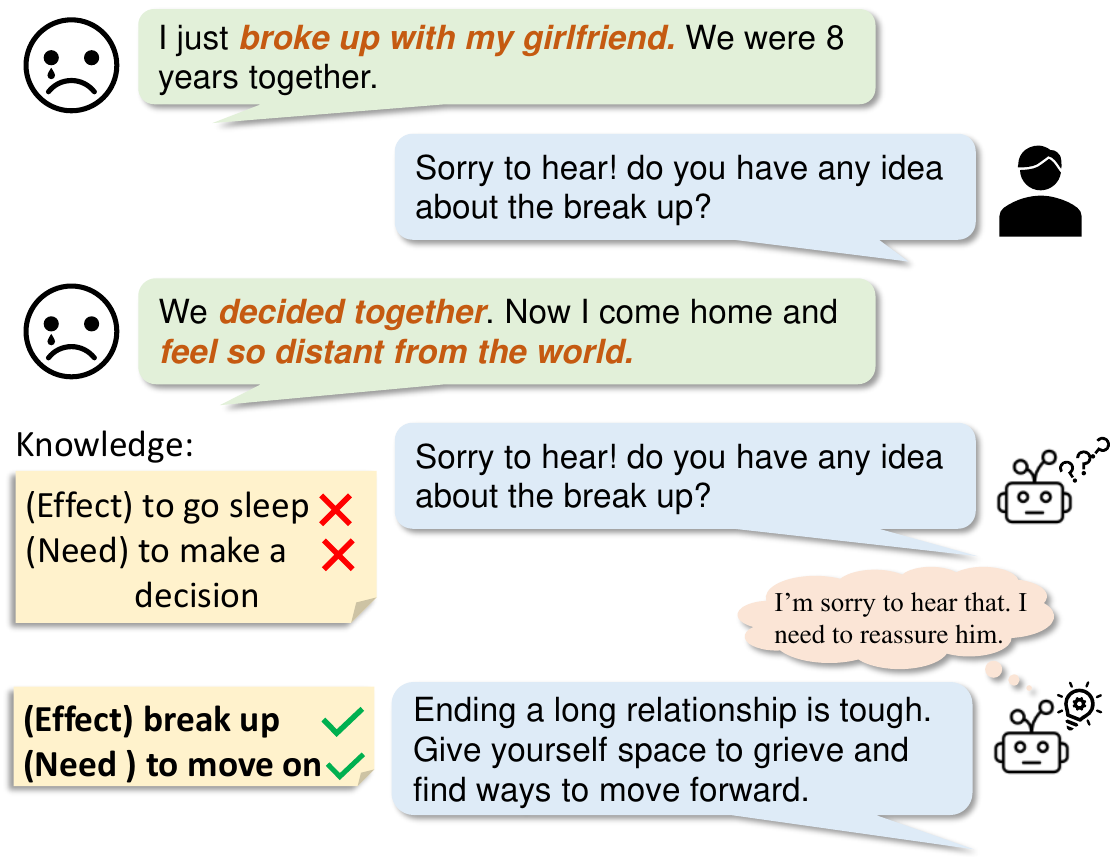}
\caption{Effect of emotional cause on empathetic response generation}
\label{figure_1_example}
\end{figure}
Existing methods elaborated various small-scale models to comprehend the speaker's emotion state based on emotional labels  \citep{majumder2020mime, tu2022misc}, or to understand the speaker's situation and experiences combined with common-sense knowledge  \citep{sabour2022cem, lin2019moel} generated by COMET  \citep{hwang2021comet}. With the rise of Large Language Models (LLMs), prompt-based methods have provided a new unified modeling approach by adding both cognition and affection information into the prompt  \citep{zhao2023chatgpt, roller2020recipes, lee2022does}. Specifically, \citet{qian2023harnessing} proposed a two-stage generation approach with few-shot Chain-of-Thought (CoT) prompt on LLMs to reason about the speaker's situation  \citep{wei2022chain}. However, these methods exhibit dependence on the language proficiency of underlying large models, leading to unstable inference performance  \citep{wei2022chain}. Furthermore, there are two extra limitations of these LLM-based models: (1) they ignore the impact of emotion cause reasoning on empathetic response generation whose importance has been proven in previous works  \citep{gao2021improving, qian2023empathetic}. (2) they lack the awareness of the role as an empathetic listeners, which makes generated responses more rational and less emotionally impactful. Figure~\ref{figure_1_example} shows an example of the empathetic listener role. 

Additionally, from the cognition aspect, directly incorporating external knowledge into large-scale models may lead to a decline in response consistency due to its lack of relevance to the contextual history of the conversation  \citep{qian2023harnessing, zhao2023chatgpt}. As shown in Figure~\ref{figure_1_example}, utilizing the irrelevant knowledge generated from the last utterance of the conversation history (e.g., "\textit{xEffect = to go sleep}" and "\textit{xNeed = to make a decision}") leads to conflicts with the correct dialogue context. Despite the efforts of DCKS  \citep{cai2023improving} which filtered knowledge according to the emotion labels to ensure emotional consistency, semantic coherence within the context remains unguaranteed. Instead, we discover that emotional causes involving further reasoning about emotions facilitates organic integration of the affection and cognition aspects and greatly alleviates the conflicts. For example, in Figure~\ref{figure_1_example}, directed by emotion causes from the dialogue history, the generated knowledge of COMET (i.e., "\textit{xEffect = break up}" and "\textit{xNeed = to move on}") successfully improves the empathetic outcomes. 

To address the above problems, we propose a novel cause-aware \textbf{C}oT \textbf{F}ine-tuning \textbf{E}mpathetic \textbf{G}eneration method (\textbf{CFEG}) on LLMs. We design an universal CoT generation template to guide the model in reflecting on its emotions and causes, and enhancing the role awareness of an empathetic listener. This is initiated by the idea that a good empathetic listener should concentrate more on understanding what triggers the emotion in the dialogue to respond for the speaker's emotion. Simultaneously, instruction tuning can be employed to strengthen the stability of the CoT reasoning. Besides, we implicitly select required common-sense knowledge from COMET directed by emotion cause to enhance the consistency between internal and external knowledge. The knowledge is then incorporated into the prompt to improve the diversity of response generation. Experimental results on the benchmark dataset demonstrate that our method, fine-tuned on LLaMA-7b, significantly outperforms existing methods, and results in more empathetic responses in human evaluations, even better than that of ChatGPT.

Our contributions are summarized as follows:

\begin{itemize}
    \item We present a novel cause-aware CoT fine-tuning generation method to enhance LLMs' capability of empathy\footnote{Our code will be publicly available.}.
    
    \item We propose to incorporate emotion cause-oriented COMET based into the CoT prompt to connect the emotional and cognitive aspects and improve the diversity of empathetic generation.

    \item We conduct extensive experiments to demonstrate the superiority of our proposed method which is able to generate more empathetic and explainable responses.
\end{itemize}

\section{Related Work}
\subsection{Empathetic Response Generation}
% 在社会心理学中，移情包括情感和认知两个角度。从情感角度出发，MIME，MISC模仿人类情感控制回复的共情效果。从认知角度出发，部分工作在模型中加入额外的认知推理步骤，（qian2023think， ），CEM借助COMET生成常识知识。 

%在社会心理学中，移情包括情感和认知两个角度。早期的工作rely on 情感信号来以模仿人类情感的方式控制回复的共情效果（MIME，MISC）。近期的工作在模型的解码中借助COMET生成额外的常识知识 for deepe congnition understanding（CEM，Knowledge Bridging）。同时，一些工作开始关注情感和认知两方面的联系。DCKS通过情感筛选外部知识，CASE通过对齐情感知识和常识知识。与这些精心设计的小模型不同，本文通过加入额外的情感归因推理步骤，借助大模型以更简单的方法连接情感和认知角度，目的在于生成更合理更相关的认知知识。
In the field of social psychology, empathy encompasses both affection and cognition aspects  \citep{davis1983measuring}. Early works relied on emotional signals to mimic human emotions  \citep{majumder2020mime, tu2022misc}. Recent works incorporated additional common-sense knowledge generated by COMET  \citep{hwang2021comet} for deeper cognitive understanding  \citep{li2022knowledge, sabour2022cem}. To enhance the relevance of incorporating knowledge with the dialogue, DCKS  \citep{cai2023improving} incorporated an emotion-based knowledge selection, while CASE  \citep{zhou2022case} aligned emotion with  knowledge. Meanwhile, \citet{qian2023think} incorporated additional reasoning steps into the model. \citet{gao2021improving, qian2023empathetic} leveraged emotion cause recognition to enhance empathetic outcomes. Different from these intricately designed smaller models, our work introduces emotion cause reasoning to generate more reasonable and relevant cognitive knowledge in a simpler manner.

\subsection{Large Language Models}
% 近年来，由于人类反馈的强化学习，大语言模型的能力取得了很大的提升。LLMs被证实有能力conduct language analysis和推理。有一些工作初步尝试将LLM应用于移情对话中。Harness通过CoT的方法探索了ChatGPT移情生成的性能。但是论文指出，CoT方法在模型参数较小时，容易引发错误的推理。随着lora微调的出现，LLM的能力进一步提高，与这个工作不同，我们的方法结合情感归因，通过微调使得相对小的模型（例如LLaMA-7b）能够更好的利用CoT，取得了比chatgpt更进一步的的推理能力和共情效果。
In recent years, the capabilities of large language models (LLMs) have significantly improved due to  reinforcement learning from human feedback (RLHF)  \citep{stiennon2020learning} and instruction tuning  \citep{ouyang2022training}. Meanwhile, LLMs can perform reasoning through the construction of prompts  \citep{liu2023gpt, pretrainpromptmpredict} and CoT  \citep{wei2022chain}. Some studies have made initial attempts to apply LLMs in empathetic dialogues. For instance, \citet{qian2023harnessing} explored the empathetic generation capabilities of ChatGPT by two-step  generation  \citep{wei2022chain}. However, \citet{wei2022chain} points out that the CoT method can lead to erroneous reasoning when the model has fewer parameters. With the advent of fine-tuning techniques based on LoRA \citep{hu2021lora}, the capabilities of LLMs have been further enhanced. Unlike the aforementioned work, our method combines emotion cause reasoning and fine-tuning, enabling relatively smaller models (e.g., LLaMA-7b) to utilize CoT more effectively, achieving superior reasoning abilities and empathetic effects beyond ChatGPT.

\subsection{Emotion Cause Pair Extraction}
The task of \textbf{E}motion-\textbf{C}ause \textbf{P}air \textbf{E}xtraction (\textbf{ECPE}) aims to recognize the emotions expressed by speakers and identify the causal spans. \citet{RECCON} provided the \texttt{RECCON} dataset for this task. \citet{gao2017overview, gui2014emotion} jointly extract emotions and causes. MGSAG \citep{MGSAG} proposed to incorporate fine-grained and coarse-grained semantic features jointly without regard to distance limitation. \citet{PRG-MoE} learned relationship between utterances and advised a gating network to incorporate dialogue features. We utilize the ECPE methods to annotate causes of the dialogue in the benchmark dataset.

\section{Problem Definition}
% 共情对话生成aims to 以倾听者的角色理解说话人的情感，并给出共情的答复。具体来说，给出n轮由说话者和倾听者交替的对话数据D = {u1,u2,...,un}，where Ui represents the i-th utterance。共情对话生成任务的目标是针对最后一句话un，识别出说话人当前的情感e，同时生成下一句具有共情效果的句子y
Empathetic generation aims to understand the emotions of the speaker in the role of a listener and provide an empathetic response. Formally, let $D = \left\{u_1,u_2,\cdots, u_n\right\}$ denotes a dialogue history with $n$ utterances, where the $i$-th utterance $u_i = \left\{w_1,w_2,\cdots, w_k\right\}$ is
a sequence of $k$ words. The goal is to identify the current emotions $e_n$ of the speaker for the last utterance $u_n$,  and play the role of the listener to generate a empathetic and informative response $Y$.

% cause-aware :ins变了，analysis the emotion and cause. P2= "ins2"这个prompt得出的结果是。。。he feels he sad result2(不要公式展示)
% listener-aware 看写在这，给出一个Result template的格式（不顺的话挪到ins里）
% {cau}输入是什么样的写清楚，COMET介绍一下输入输出是什么。
% 知识也要加在prompt中，怎么加的得到又一个prompt
% 现在有了前面的prompt，现在需要tuning，需要构造一个prompt+result的pair，进一步，ins tuning里可以利用ICL，约束输出的格式，finetuning时给出P3'，response template不变。

\section{Method}
% In this section, we provide a detailed introduction to the \textbf{CFEG} method. We first present the Cause-Aware prompt construction and introduce the knowledge generation method. Then, we introduce the listener-aware response strategy and fine-tuning process.
Our proposed \textbf{CFEG} method is basically fine-tuned on LLMs. In this section, we first present how to construct cause-aware CoT prompt template consisting of the instruction, dialogue context, and external knowledge. Then, we introduce the CoT output template and the fine-tuning method.
% common prompt

\subsection{Cause-Aware Prompt Construction}
\noindent\textbf{Common Prompt}\\
LLMs have been demonstrated to perform empathetic generation given appropriate task instructions  \citep{roller2020recipes, lee2022does}. Given the dialogue history $D$ as the input, one of the basic prompt for this task is as follows:
\begin{equation}
\begin{split}
% &P_1 = "\text{Analyze emotion and respond with}\\ 
% &\text{empathy to the provided dialogue:} \{D\}
&P_1 = "\{Ins_1\}. \text{The Dialogue: } \{D\}."
\end{split}
\end{equation}
Here, $\{Ins_1\}$ represents the task instruction: \textit{"Analyze emotion and respond empathetically to the provided dialogue"}. $\{D\}$ represents the dialogue history which consists of the roles (speaker, listener) and the dialogue utterances, and we use ";" to concatenate multiple turns. Then, we input the prompt to a LLM, e.g., LLaMA-7b  \citep{touvron2023LLaMA} to generate the empathetic response $\{Res\}$. Typically, the responses may take the form of "\textit{He feels ..., I will reply as follows: ...}". %Through the investigation of Cause-Aware strategy, we further guide the model to infer the user's emotions and reasons, thereby enhancing the empathetic effect.

\noindent\textbf{Cause-Aware CoT Strategy}\\
% Inspired by \citet{RECCON}. we propose guiding the model to extract emotional causes from the conversation history through a designed cause-aware prompt. Previous works  \citep{qian2023harnessing,zhao2023chatgpt} used CoT to help LLMs infer the speaker's situation. However, these approaches did not effectively leverage causal information in the dialogue history, potentially causing a deviation between inference and dialogue  \citep{wei2022chain}. Therefore, we construct a generation template to enhance stability through fine-tuning on the dataset. The template is as follows:
%Inspired by \citet{RECCON}. we propose a cause-aware CoT prompt to guide the model in extracting emotional causes (formalized as a phrase encompassing start and end positions) from the dialogue history. 
CoT refers to a series of intermediate reasoning steps  \citep{brown2020language}, which controls the direction of the model's thinking through multiple steps. In the empathetic generation task, LLMs are required to infer the speaker's emotion and the situation. Therefore, previous works  \citep{qian2023harnessing} utilized CoT to generate an appropriate and more human-like response. Specifically, in the first stage, the model is prompted with "\textit{Don't rush to reply yet, what may be the user's emotion, and what may be the situation?}" to guide speculation on the situation based on the user's statement. Then, in the second stage, prompted with "\textit{Combine your thoughts with the dialogue context and give your response.}", the final response is generated. However, this kind of CoT methods relies on the model's linguistic capabilities, leading to uncontrollable situation reasoning that may be inconsistent with the dialogue history  \citep{wei2022chain}. 

Different from directly inferring the situation \citep{qian2023harnessing}, we propose to use a cause-aware CoT strategy to guide the model in first extracting emotional causes (formalized as a phrase encompassing start and end positions), and then generate responses. Our CoT prompt is as follows:
%identify causes within the history to constrain the model's reasoning, while also leveraging fine-tuning through response template construction to further enhance reasoning coherence.
%因此我们从对话历史中寻找原因，以约束模型的推理，同时可以通过后续构造回复模板借助微调进一步增强推理的合理性。
% Therefore, we construct a generation template to enhance stability through fine-tuning on the dataset. The template is as follows:
\begin{equation}
\begin{split}
P_2= "\{Ins_2\}. \text{ The Dialogue: } \{D\}."
% &P_2 = \text{Analyze the emotion and identify}\\
% &\text{the cause from the dialogue. Then respond}\\
% &\text{with empathy to the provided dialogue:}\{D\}
\end{split}
\end{equation}
where $\{Ins_2\}$ represents the cause-aware task instruction:"\textit{Analysis the emotion and identify the cause from the dialogue. Then respond empathetically to the provided dialogue.}". Results are generated from LLMs in the form of \textit{"He feels ... because he says .... I will reply as follows: ..."}.
% \begin{equation}
% \begin{split}
% P_2=\text{He feels} \{emo\} \text{because he }\text{says} \{cau\}
% \end{split}
% \end{equation}
% Here $\{emo\}$ represents the speaker's emotion and $\{cau\}$ represents the speaker's causal span in the history that needs to be predicted.

\subsection{Cause-Oriented COMET}
% COMET  \citep{bosselut2019comet} to generate commonsense knowledge, which is a generative model pre-trained on the commonsense knowledge base $\text{ATOMIC}_{20}^{20}$  \citep{hwang2021comet}.
COMET  \citep{hwang2021comet} is a pre-trained GPT-2 model  \citep{radford2018improving} that has been fine-tuned on triplets (\textit{e, r, i}) extracted from the ATOMIC dataset  \citep{hwang2021comet}, where \textit{e} represents the event, \textit{r} represents the relation type, and \textit{i} represents the inferred knowledge. Five common-sense relations of inferences are generated: the impact of events on individuals (\textit{xEffect}), their reactions to events (\textit{xReact}), the intentions prior to events (\textit{xIntent}), the requirements for events to occur (\textit{xNeed}), and the desires following events (\textit{xWant}). 

Previous methods  \citep{sabour2022cem, zhou2022case,qian2023harnessing} utilize COMET to acquire external knowledge from the last turn of the dialogue, which is then fused into the model to improve the diversity of generation. However, we find that there exist conflicts between the external knowledge and the dialogue context. To settle this problem, we propose to use COMET knowledge generated from the emotional cause-span instead: 

\begin{equation}
kg_i^{ecpe} = COMET(r_i,\{cau\})  
\end{equation}
% \begin{equation}
% P_3^{ecpe} =  \oplus_{i=1}^n \{kg_i^{ecpe}\} 
% \end{equation}
where $r_i\in$ \{\textit{xReact, xWant, xNeed, xIntent, xEffect}\},  and $\{cau\}$ represents the causal span extracted from the history (e.g., in Figure~\ref{figure_1_example}'s dialogue, the causal span is '\textit{I just broke up with my girlfriend.}'). We concatenate the knowledge and transform it into natural language segments $\oplus_{i=1}^5 kg_i^{ecpe}$ (e.g., \textit{"He tends to look nice; He needs to have a haircut; He wants to fix his hair; The effect is that he ends up burning his hair; He feels embarrassed."}). 

We then incorporate the cause-oriented COMET knowledge into the CoT prompt:
\begin{equation}
\begin{split}
P_2^{kg}= &``\{Ins_2\}. \text{The Dialogue: } \{D\}. \text{ In this}\\ &\text{Dialogue, }\{\oplus_{i=1}^5 kg_i^{ecpe}\}."
\end{split}
\end{equation}
\subsection{Instruction Tuning}
% 现在有了前面的prompt，现在需要tuning，需要构造一个prompt+result的pair，进一步，ins tuning里可以利用ICL，约束输出的格式，finetuning时给出P3'，response template不变。
\noindent\textbf{Output Template} \\
Instruct tuning is employed to further enhance the model's empathetic expression. As the format of output plays an important role in the fine-tuning procedure, we also wrap the emotion reasoning and response into a natural language template. Normally, the output format of common prompt $P_1$ is as follows:
\begin{equation}
\begin{split}
R_1 = &``\text{He feels} \{emo\}.\\ &
\text{I will reply him:\{response\}}."
\end{split}
\end{equation}

We design the cause-aware CoT output format of $P_2$ as follows:
\begin{equation}
\begin{split}
R_2 = &``\text{He feels} \{emo\} \text{because he } \text{says} \{cau\}. \\ &
\text{I will reply him:\{response\}}."
\end{split}
\end{equation}
Here $\{emo\}$ represents the speaker's emotion and $\{cau\}$ represents the speaker's causal span in the history that needs to be predicted.

\noindent\textbf{Listener-Aware CoT Strategy}\\ Empathy is inherently subjective, influenced by both the speaker's description and the listener's feeling. Therefore, transforming the output template into a listener-aware format helps the model differentiate the speaker's emotions from its own responsive emotions. Different from previous work  \citep{zhao2022don}, which solely perceived the emotions of the speaker and listener, we have further enhanced the model's capability to infer conversational intent. The listener-aware CoT output template is as follows:
\begin{equation}
% \centering
\begin{split}
R_2^{la} = &``\text{He feels} \{emo\} \text{because he } \text{says} \\
&\{cau\}.  \text{ I'm } \{emo\} \text{to hear that. } \\ & \text{I will} \{Intend\} \text{him:\{response\}}."
\end{split}
\end{equation}
Here, $\{emo\}$ represents the listener's emotion reflected by the model. The model chooses "\textit{glad}" or "\textit{sorry}" based on the user's emotion, and $\{Intend\}$ represents the conversational intent, selecting either “\textit{reassurance}” or “\textit{sympathize}” based on the user's emotion.

\noindent\textbf{Demonstration}\\
LLMs possess the ability of in-context learning(ICL)  \citep{brown2020language}, a small amount of data examples can enhance the performance of the model. Inspired by \citet{qian2023harnessing}, given the current dialogue, we sample 5 complete dialogues containing replies from the training dataset to construct the demonstration in the format:
% \begin{equation}
% % \centering
% \begin{split}
% E_i = & \{D_i\} \text{. In this Dialogue,} \{P_{3i}^{ecpe}\} \\
% & \text{SYS:} \{R_1\} \text{. } \{R_2\} \text{. }\{Res_i\}
% \end{split}
% \end{equation}
% Where $\{Res_i\}$ represents the true responses in the dataset. We enrich the input of the large model by incorporating common knowledge and dialogue examples, constructing the following final prompt template: 
\begin{equation}
\centering
\begin{split}
E = &``\text{I'll give you five examples.}
\\ & Examples\{D_1,...D_5\}."
\end{split}
\end{equation}

\noindent\textbf{Loss Function}\\
The demonstration is added after $\{Ins2\}$ in $P_2^{kg}$ to get our final input prompt $P_2^{kg+E}$. 
%我们构造cause-aware的CoT generation template。在共情回复前分析speaker的情感和原因，并且生成lisener的情感和共情意图。 The reply template is as follow:
After the prompt and output template are designed, we transfer all the samples in the datset into a prompt and output pair $<P_2^{kg+E},R_2^{la}>$.
% We construct the cause-aware CoT generation template to guide the LLMs to analyze the speaker's emotions and causes before generating empathetic responses, facilitating the generation of the listener's emotions and empathetic intentions. The reply template is as follow:
% \begin{equation}
%     Y = \{R_1\}.\ \{R_2\}.\ \{Res\}.
% \end{equation}
% The objective of the model is to generate emotion cause reasoning and empathetic responses (e.g.,\textit{"He feels sad because he says 'I just broke up with my girlfriend'. I'm sorry to hear that. I will reassurance him: ..."}). 
The supervised fine-tuning loss of the LLMs is as follows:
\begin{equation}
    \mathcal{L} = -\sum_{j} \log p_{\phi}(R_t|P,R_{<t})
\end{equation}

\section{Experiment}
\subsection{Experiment Setup}
% 我们在EmpatheticDialog数据集上进行实验。数据集包含25k个对话，每个对话包含两轮，其中说话者倾诉个人经历，听者推断说话者的处境和情绪，并以同理心做出回应。所有对话are labeled with $32$ emotion categories.我们按照官方将训练集，验证集，测试集以8:1:1进行划分。

\noindent\textbf{Dataset} We conduct experiments on the \texttt{EmpatheticDialogue}  \citep{rashkin2018towards}. The dataset comprises $24,850$ dialogues, each annotated with one of $32$ emotion categories, and involved two turns of empathetic conversation between a speaker and a listener. Following previous works  \citep{rashkin2018towards}, we randomly split the train/valid/test sets in an $8:1:1$ ratio.

% \noindent\textbf{Dataset} We conduct experiments on the \texttt{EmpatheticDialogue} benchmark dataset  \citep{rashkin2018towards}. The dataset comprises $24,850$ dialogues, each with two turns: a speaker sharing personal experiences and a listener responding empathetically. Each dialogue is annotated with one of $32$ emotion categories. Following \citet{rashkin2018towards}, we randomly split the train/valid/test sets in an $8:1:1$ ratio.
\begin{table*}[ht]
\centering
\resizebox{\textwidth}{!}{%
\begin{tabular}{@{}lcccccc|cccc@{}}
\toprule
 \textbf{Model}  & $Acc$ & $PPL$ & $Dist$-$1$ & $ Dist$-$2$ & $BLEU$-$2$ & $BLEU$-$4$ & $Coh.$ & $Emp.$ & $Inf.$ & $Flu.$\\ 
\midrule
MIME & 30.96 & 37.24 & 0.47 & 1.66 & 6.78 & 1.94 & - & - & - & - \\
CEM & 36.84 & 36.33 & 0.62 & 2.39 & 5.64 & 1.70 & 3.08 & 2.89 & 3.02 & 3.95 \\
DCKS & 49.16 & 16.08 & 2.19 & 9.61 & - & - & 3.07 & 3.21 & 3.31 & 4.05 \\
EmpSOA & 48.32 & 35.02 & 0.71 & 3.96 & - & -  & - & - & - & - \\
CASE & 40.2 & 35.37 & 0.74 & 4.01 & 7.10 & 2.27 & 3.01 & 3.34 & 2.87 & 4.11 \\ 
LLaMA-7b+ICL & 45.53 & 13.26 & 2.07 & 8.48 & 4.14 & 1.41 & 3.41 & 3.64 & 3.22 & 4.12 \\
LLaMA-7b+CoT & 46.44 & 11.42 &  2.09 & 9.19 & 4.11 & 1.36 & 3.32 & 3.70 & 3.23 & 4.09 \\
LLaMA-7b+CKG & 46.12 & 10.35 & 2.11& 9.76 & 4.05 & 1.09 & 3.39 & 3.59 & 3.62 & 4.14 \\
LLaMA-7b+$kg^{ecpe}$ & 47.31 & 10.65 & 2.32& 10.06 & 4.59 & 1.59 & 3.57 & 3.66 & 3.79 & 4.13 \\
ChatGPT & 47.42 & 6.79 & 2.72 & 17.12 & 6.19 & 1.86 & 4.15 &4.12 &4.01 &\textbf{4.78} \\
ChatGPT+ICL & 48.26 & 5.68 & 2.51 & 16.09 & 7.07 & 2.23 & 4.19 & 4.22 & 4.07 & 4.72 \\
ChatGPT+CoT & 48.74 & 5.31 & 2.48 & 16.90 & 4.99 & 1.37 & 4.23 & 4.30 & 4.12 & 4.65 \\
ChatGPT+CKG & 48.18 & \textbf{5.26} & 2.73 & 18.29 & 5.31 & 1.47 & 4.16 & 4.15 & 4.16 & 4.70 \\
ChatGPT+$kg^{ecpe}$ & 48.66 & 5.01 & 3.01 & 19.24 & 7.59 & 2.51 & 4.26 & 4.27 & 4.43 & 4.71 \\
\midrule
CFEG & \textbf{52.73} & 6.67 & \textbf{2.96} & \textbf{19.52} & \textbf{10.54} & \textbf{5.17} & \textbf{4.32} & \textbf{4.51} & \textbf{4.51} & 4.49\\ 
w/o $R_2$ & - & 9.31 & 2.23 & 15.17 & 7.12 & 4.81 & 3.61 & 3.76 & 3.47 & 4.24 \\ 
w/o $R_2^{la}$ & 49.77 & 9.89 & 2.33 & 14.13 & 7.85 & 3.15 & 3.94 & 4.01 & 3.87 & 4.77 \\ 
w/o $kg^{ecpe}$ & 45.91 & 10.20 & 2.15 & 14.94 & 9.14 & 4.62 & 3.78 & 3.92 & 3.27 & 4.15 \\ 
w/o $E$   & 46.50  & 10.65 & 2.24 & 11.20  & 8.33 & 3.26 & 3.78 & 3.92 & 3.27 & 4.15 \\ 
w/o $sft$ & 46.23 & 11.72 & 2.35 & 10.11 & 3.25 & 2.56 & 3.27 & 3.65 & 3.21 & 4.08 \\  
\bottomrule
\end{tabular}}
\caption{The automatic and human evaluation results between our model and the baseline, as well as the results for our model ablation experiments, respectively. The results of our CFEG method are averaged across 5 experiments.\label{tab_1_main}}
\end{table*}
% 实验式通过几次，random，什么样的参数，取平均
 
% \caption{The results of human ratings for our model, as well as competitive baselines, and the results of our model ablation experiments (significance tested with a t-test with a p-value < 0.05).\label{tab: human rating}}

\noindent\textbf{Emotion Cause Annotation} As our method requires additional emotional cause-spans, we train a LLaMA-7b model on the \texttt{RECCON} dataset \citep{RECCON}, which is utilized for conversation emotional cause-span recognition. The model achieves a $macro\_F_1$ score of $74.16\%$ on the test set. We then utilize this model to make inference on the \texttt{EmpatheticDialogue} dataset. Manual evaluation of 100 randomly sampled dialogues results in a  $macro\_F_1$ score of $72.34\%$ on the \texttt{EmpatheticDialogue} dateset, demonstrating  sufficient performance for conducting CoT reasoning and generating external knowledge in our method.

\noindent\textbf{Evaluation Metrics} 
We assess the models' performance using both automatic and human evaluations. For automatic evaluation, we employ Perplexity (\textbf{PPL}) for generation quality, Distinct-n (\textbf{Dist-1/2})  \citep{li2015diversity} for response diversity, BLEU-n (\textbf{BLEU-2/4})  \citep{papineni2002bleu} for response similarity and relevance, and emotion accuracy (\textbf{ACC}) for emotion prediction. Human evaluation intuitively validates the model's expression and empathy which consist of on four aspects: \textbf{Coherence (Coh.)}, assessing relevance to the context; \textbf{Empathy (Emp.)}, evaluating understanding and empathetic expression; \textbf{Informative (Inf.)}, measuring incorporation of external knowledge; and \textbf{Fluency (Flu.)}, assessing naturalness. During the evaluation process, we randomly select 200 conversation contexts. Annotators consist of both graduate students and experienced experts who have undergone systematic training. They are asked to score each response on a scale from 1 to 5(1: not at all, 3: OK, 5: very good).

\noindent\textbf{Baselines } We compare our methods with both existing small-scale models and LLMs.
\begin{itemize}
%（1）MoEL在decoder中融合多种情感；（2）MIME通过模仿人类情感进行回复；（3）MISC：使用情感支持混合策略并结合知识进行情感支持生成；（4）CEM：融入了额外的外部常识知识；（5）KEMP：使用concept知识构建情感语境图提升共情能力（5）DCKS： incorporated an emotion-based adaptive module for commonsense knowledge selection;(6)EmpSOA:generate Empathetic response with explicit Self-Other Awareness；(7)CASE: 从粗粒度和细粒度层次上对齐常识认知图和情感概念图。
% \item \noindent\textbf{smaller scale model:}  (1)\textbf{MoEL} integrated multiple emotions in the decoder  \citep{lin2019moel};(2) \textbf{MIME} generated responses by imitating human emotions  \citep{majumder2020mime}; (3) \textbf{MISC} employed an emotion-supportive hybrid strategy combined with knowledge for empathetic generation  \citep{tu2022misc}; (4) \textbf{CEM} incorporated additional external commonsense knowledge  \citep{sabour2022cem}; (5) \textbf{KEMP} enhanced empathetic capabilities by building an emotional context graph using concept knowledge  \citep{li2022knowledge}; (6) \textbf{DCKS} incorporated an emotion-based adaptive module for commonsense knowledge selection  \citep{cai2023improving}; (7) \textbf{EmpSOA} generated empathetic responses with explicit Self-Other Awareness  \citep{zhao2022don}; (8)\textbf{CASE} aligned commonsense cognitive graphs and emotional concept graphs at both coarse-grained and fine-grained levels.

\item \noindent\textbf{Non-LLMs:}  (1)\textbf{MIME} generated responses by imitating human emotions  \citep{majumder2020mime}; (2) \textbf{CEM} incorporated additional external commonsense knowledge  \citep{sabour2022cem}; (3) \textbf{DCKS} incorporated an adaptive module for common-sense knowledge selection  \citep{cai2023improving}; (4) \textbf{EmpSOA} generated empathetic responses with  self-other awareness  \citep{zhao2022don}; (5)\textbf{CASE} aligned  cognitive and emotional  graphs \citep{zhou2022case}.

%CoT工作使用大模型生成共情对话
%参考harness的工作，chatgpt+ICL 方法在prompt中加入 Semantically Similar In-Context Learning，Chatgpt+knowledge 方法通过comet加入常识知识
\item \noindent\textbf{LLMs:} 
\textbf{LLaMA-7b}  \citep{touvron2023LLaMA} and \textbf{ChatGPT} have been chosen as the baseline models for empathetic generation. Following \citet{qian2023harnessing}, we add the following strategies to these two models: (1) \textbf{+ICL} involved incorporating semantically similar In-Context Learning  \citep{liu2021makes}; (2) \textbf{+CoT} inferenced the speaker's situation before response; (3) \textbf{+CKG} utilized the last utterance of the history to integrate commonsense knowledge by COMET  \citep{hwang2021comet}. Meanwhile, \textbf{+$kg^{ecpe}s$} utilize the cause-oriented COMET to generate higher-quality knowledge, distinguishing it from previous methods  \citep{qian2023harnessing}. 
\end{itemize}

\begin{table}[t]
\centering
\resizebox{\linewidth}{!}{
\begin{tabular}{@{}cccc@{}}
\toprule
Comparisons                   & Aspects & Win & Lose \\ \midrule
\multirow{3}{*}{CFEG vs. CASE} 
& Coh.  & 87.5\%   & 5.5\%     \\
& Emp.	& 94.5\%   & 3.0\%     \\
& Inf.  & 88.0\%   & 4.0\%     \\ \midrule
\multirow{3}{*}{CFEG vs. LLaMA-7b} 
& Coh.  & 66.0\%   & 15.5\%    \\
& Emp.  & 71.0\%   & 14.5 \%   \\
& Inf.  & 76.5\%   & 11.5\%    \\ \midrule
\multirow{3}{*}{CFEG vs. ChatGPT} 
& Coh.  & 59.5\%   & 36.5\%    \\
& Emp.  & 59.0\%   & 37.5\%    \\
& Inf.  & 54.0\%   & 34.0\%    \\ \bottomrule
\multirow{3}{*}{\shortstack{CFEG \\ vs. \\ ChatGPT+CoT}} 
& Coh.  & 54.0\%   & 38.0\%    \\
& Emp.  & 55.0\%   & 38.5\%    \\
& Inf.  & 53.5\%   &37.0\%    \\ \bottomrule
\end{tabular}}
\caption{The result of human A/B test on aspects compared with baselines.\label{tab_2_human_ABtest}}
\end{table}

\begin{table*}[t] % 't' 表示表格置于顶部
\centering % 居中显示表格
\begin{tabularx}{\textwidth}{@{}p{1.9cm}|X@{}} % 使用 tabularx 环境
\toprule
Emotion & Embarrassed \\ 
\midrule
Context & Speaker: I burned my hair with my hair dryer. I am so embarrassed to go out. \\
         & Listener: I am sorry to hear that, my wife has done the same thing. She wore a hat for almost a month when she left the house. \\
         & Speaker: Yeah, it is the worst, I look so weird with my hair like this, it is so embarrassing. \\
         & Ground-Truth: Well, like I told her, it will grow out, it will just take time. \\ \midrule
Knowledge
         & COMET: Intent to look good; Need to have a hair dryer; Want to wash their hair; Effect to get a haircut; feels embarrassed \\
         & Cause-oriented COMET: Intent to look nice; Need to have a haircut; Want to fix the hair; Effect: burn hair; feels embarrassed \\ 
\midrule
MIME & That is a pretty funny story. \\
CEM & I am sorry to hear that. \\
CASE & I hope you are able to get it fixed, or just let her know it is not too bad. \\
\midrule
ChatGPT & I totally understand how embarrassing it can feel, but remember, accidents happen to all of us at some point. \\ \cline{2-2}
+Knowledge & I understand how upsetting it can be to deal with a hair mishap. Remember, it's a temporary situation. How about trying some new hairstyles? \\ \cline{2-2}
+Cause-oriented Knowlegde & It sounds like you've had a really tough experience with your hair dryer, and it's completely understandable to feel embarrassed about how your hair looks right now. These kinds of accidents can happen to anyone, and it will grow back. \\
\midrule
CFEG & Prompt: Analysis the emotion and identify the cause from the dialogue. Then respond empathetically to the provided dialogue. I'll give you five examples. Examples: ... The dialogue: ... .In this Dialogue: He tends to look nice; He needs to have a haircut; He wants to fix his hair; The effect is that he ends up burning his hair; He feels embarrassed.

SYS:A feels embarrassed because he says "I burned my hair with my hair dryer." I feel so sorry to hear that. Therefore, I will reassure him:\textbf{ I understand how you feel, but remember it's just temporary.Your hair will grow back.Perhaps get a haircut to fix it.} \\
\bottomrule
\end{tabularx}
\caption{Generated responses from baselines and our model. The general knowledge generated by COMET and ECPE oriented COMET is given in the Context row. \label{tab_3_case}}
\end{table*}

\noindent\textbf{Implementation Details}
The overall project framework is implemented using \emph{ LLaMA-Factory}\footnote{\url{https://github.com/hiyouga/LLaMA-Factory}}. The LLaMA-7b model is downloaded from the open-source  \emph{huggingface Transformers}\footnote{\url{https://github.com/huggingface/transformers}}. We perform fine-tuning on the model using LoRA \citep{hu2021lora}, with a learning rate set to $5e^{-5}$, LoRA rank 8, and a batch size of 4. The common-sense knowledge is generated from COMET\footnote{\url{https://github.com/allenai/comet-atomic-2020}}. In order to ensure deterministic outputs in our experiments, we set the temperature to 0. We use the model gpt-3.5-turbo provided in the OpenAI API for the baselines, which is the base model of ChatGPT\footnote{\url{https://platform.openai.com/docs/models/gpt-3-5-turbo}}. The training is conducted on a server equipped with 8 NVIDIA RTX 3090 GPUs, utilizing the Accelerate framework\footnote{\url{https://github.com/huggingface/accelerate}}.All experiments are conducted on 5 random seeds. We select the model with best performance on the validation set and run it on the test set to report its average results.

\subsection{Experimental Results}
\subsubsection{\textbf{Automatic Evaluation}}
Table~\ref{tab_1_main} shows the main results of our method and other baselines. Experiment results demonstrate that our CFEG method achieves the best performance on all metrics except the $PPL$ metric, indicating that incorporating CoT fine-tuning can enhance the model's emotion understanding capability. Specifically, compared to all the non-LLMs models, the LLMs-based approaches exhibit significant advantages due to the inherent linguistic capabilities of the model itself. In comparison with ChatGPT+CoT, the CFEG method improves emotion accuracy by $4.53\%$ while also enhancing $BLEU$-$1$/$2$ scores by $5.55\%$ and $3.80\%$, indicating that our cause-aware CoT strategy guides the model to analyze emotions and causes from the history, resulting in better human-like empathetic outcomes. Meanwhile, Providing common-sense knowledge leads to improvements in both LLaMA-7b and ChatGPT on the $Dist$-$2$/$4$ metrics. However, compared to ChatGPT+CKG, the CFEG method shows a further improvement of $0.23\%$ and $1.23\%$, respectively. attributed to the higher-quality knowledge generated by cause-oriented COMET aligning better with the context. It's worth mentioning that providing cause-oriented knowledge to LLaMA and ChatGPT can also further enhance empathetic effectiveness.
% Experiment results demonstrate that our CFEG method achieves the best performance on all metrics except the PPL value. Specifically, compared to all the non-LLMs models, the LLMs-based approaches exhibit significant advantages due to the inherent linguistic capabilities of the model itself. In comparison with ChatGPT+CoT, the CFEG method improves the accuracy of emotion by $4.53\%$, indicating that incorporating CoT fine-tuning can enhance the model's emotion understanding capability. Meanwhile, compared to ChatGPT+$\{P_4\}$, the CFEG method demonstrates a $2.95\%$ and $2.66\%$ improvement in BLEU-1/2, indicating that our method can make responses more consistent with those of real speakers. Compared to ChatGPT+CKG, the CFEG method shows a $0.23\%$ and $1.23\%$ improvement in Dist-2/4 metrics, attributed to the ECPE-oriented COMET providing higher-quality external knowledge. By comparing LLaMA and ChatGPT models with different external knowledge (+CKG and +$\{P_4\}$), it is also observed that ECPE-oriented COMET aligns better with the context, resulting in improvements in BLEU and Dist metrics for both models.

\subsection{Human Evaluation}
The human-evaluated results shown in Table~\ref{tab_1_main} demonstrate that our CFEG method outperforms the baseline in the $Coh.$, $Emp.$, and $Inf.$ aspects. Particularly, the superiority of our cause-aware CoT finetuning method in empathy and informativeness indicates its advantage in cognitive empathy and affective interaction, supporting the observations from automatic evaluations. The $Flu$. score of the CFEG method is inferior to ChatGPT, mainly because we utilize LLaMA-7b for response generation, which has significantly fewer parameters compared to ChatGPT, resulting in a gap in language capability. Additionally, Providing cause guided  knowledge to ChatGPT leads to improvements in both $Coh.$ and $Inf.$ score, underscoring the superiority of our knowledge generation method.

Meanwhile, following \citet{sabour2022cem}, we conduct an aspect-based pairwise preference test where annotators choose the better response from two results. The results are listed in table ~\ref{tab_2_human_ABtest}. We observe that our model also outperforms all the baselines, which confirmed that our method can improve the empathy effect of responses. Compared to ChatGPT+CoT, it can be seen that in $55\%$ of cases in the A/B test, human annotators prefer responses generated by the CFEG model. This indicates that our CoT fine-tuning method enable better understanding of user affection and cognition.
% When ChatGPT is provided with cause-guided external knowledge, our CTEG method still exhibits better empathy effects in $53\%$ of cases, along with improved contextual coherence. This is because, under the same external knowledge, our cause-aware prompt construction better guides the large model in emotional cognitive reasoning. 

\subsection{Ablation Study}
To analyze the performance of different strategies, we conduct experiments on the following modifications: (1) \textbf{w/o $R_i$}: We conduct ablation studies on individual prompts to observe the influence of causal reasoning and listener-aware reasoning. (2) \textbf{w/o $kg^{ecpe}$}: We remove the  external knowledge to observe its impact on empathetic generation; (3) \textbf{w/o $E$}: We  ablate examples to observe the impact of in-context learning; (4) \textbf{w/o $sft$}: We perform  ablation on fine-tuning to observe the effect of fine-tuning CoT reasoning.

As shown in Table~\ref{tab_1_main}, the model with all modules exhibited better performance. Firstly, when $R_2$ is removed, there is a decrease of $3.42\%$ and $0.36\%$ in $BLEU$-$2$/$4$ scores, and in manual analysis, the $Emp.$ metric shows the most significant decrease. This is because the model lacks reasoning about emotions and causes. Secondly, when $R_2^{la}$ is removed, both automatic evaluation metrics and manual evaluation decrease, indicating that listener-aware reasoning is closer to the conscious process of expressing empathy in humans. Additionally, when $kg^{ecpe}$ is removed, the $Inf.$ metric decreases by $0.94$, indicating that external knowledge can effectively enhance the informativeness of responses. Removing $E$ weakens the patterns learned by the model from instructions. Performance is lowest when no fine-tuning is performed. Fine-tuning ensures the stability of CoT reasoning while learning genuine human expressions.

\begin{table}[t]
\centering
\begin{tabular}{@{}lcc@{}}
\toprule
\textbf{Method}                      & $Emp.$ & $Coh.$ \\ \midrule
COMET+last utterance       & 3.61  & 3.24  \\
COMET+selection(DCKS)                       & 4.15  & 3.98  \\
ChatGPT-generated knowledge                   & 4.00  & 3.85  \\
Cause-oriented COMET(CFEG) & \textbf{4.12}  & \textbf{4.47}  \\ \bottomrule
\end{tabular}
\caption{The results of human ratings for generations on different external knowledge. The inter-annotator agreement achieved a kappa coefficient of $79.8\%$. \label{tab_4_kg}}
\end{table}
\subsection{Case Study}
% 可以看出，CFEG模型更可能以高度移情的语气回复，并与对话历史更加一致。这得益于我们方法的两大优势：一方面，相比直接使用COMET，cause-oriented COMET提供了更加准确的外部知识，减少了模型的误解。在加入ECCOMET后，ChatGPT的共情回复也会进一步提升。另一方面，我们在回复时通过思维链推理从对话历史中寻找情感原因，并且反思作为listener的情感，这样的推理兼备情感理解和认知理解，对于共情回复帮助更大。
The generated responses from our method and the compared baselines are list in Table~\ref{tab_3_case}. Our CFEG model is more likely to respond in a highly empathetic tone and is more consistent with the conversation. This is attributed to two major advantages: on one hand, cause-oriented COMET provides high-quality common-sense knowledge, reducing the model's misunderstandings. ChatGPT also further enhances empathetic effects with cause-oriented knowledge. On the other hand, we utilize CoT reasoning during responses to search for emotional causes, and reflect on the emotions as a listener, such reasoning combines affective and cognitive understanding, which aids in empathetic responses. More cases can be found in Appendix A.
%可以看出，当给出情感原因对时，COMET生成的外部知识与对话历史的相关性明显提升，这是的对话回复中包含更多的用户处境的理解。当通过提示模板让模型具有倾听者的意识后，模型的回复会更加关注理解情感和处境后的自我感受，从而更具有情绪上的感染力。

\section{Discussion}
\subsection{Effect of External Knowledge}
% 在这一节中我们进一步分析外部知识的生成质量。harness provided the last dialogue turn to COMET。基于此，DCKS利用情感对知识进行排序。与他们不同，我们利用对话历史中的情感原因span生成知识。最后我们也利用ChatGPT直接生成外部知识。
In this section, we further analyze the quality of knowledge generated by different methods, focusing on two perspectives: emotional consistency and contextual coherence. \citet{qian2023harnessing,sabour2022cem} provided the last dialogue turn to COMET. DCKS  \citep{cai2023improving} selected knowledge using emotions. In contrast, Our CFEG method generate knowledge oriented by causal spans. Finally, we also utilize ChatGPT to directly generate knowledge. Specifically, three evaluators rate them on a scale of 1 to 5 for Empathy and Coherence (e.g., in Table~\ref{tab_3_case}, while the emotion tone of the knowledge is consistent, "\textit{Need to have a hair dryer}" conflicts with the history, hence receiving a Coherence score of 3 and an Empathy score of 5). The experimental results are shown in Table~\ref{tab_4_kg}. It can be observed that, compared to DCKS and ChatGPT, the CFEG method achieves higher consistency scores. This is due to the incorporation of causal information, giving us an advantage in maintaining consistency within the dialogue history.

\begin{table}[t]
\resizebox{\linewidth}{!}{%
{\fontsize{30}{40}\selectfont
\begin{tabular}{@{}lccc@{}}
\toprule
\textbf{Template}     &$Acc$ & $F_1$   & $Emp.$  \\ \midrule
Emotion: \{emo\} Cause: \{cau\}.    & 49.61    & 62.10    & 4.02  \\
$\text{He says} \{cau\} \text{and he feels} \{emo\}.$  & 51.48    & 63.78    & 4.27 \\
$\{cau\} \text{makes him feel} \{emo\}. $ & 50.43    & 63.58 & 4.13\\
$\text{He feels} \{emo\} \text{upon saying} \{cau\}.$  &51.55  &64.95 & 4.32  \\
$\text{He feels} \{emo\} \text{because he says} \{cau\}.$  & \textbf{52.73} & \textbf{67.27} &\textbf{4.51}   \\ 
\bottomrule
\end{tabular}}}
\caption{The effect of CoT output templates $R_2^{la}$.\label{tab_5_template}}
\end{table}

% The experimental results are shown in Table 5. We run two baselines proposed by \citet{qian2023harnessing, cai2023improving}, alongside utilizing ChatGPT for knowledge generation to compare with the CFEG method in this experiment. DCKS  \citep{cai2023improving} ranks knowledge based on emotion, hence achieving good emotional consistency; however, due to the incorporation of causal information, we have an advantage in dialogue history consistency. CFEG also outperforms ChatGPT, primarily because COMET exhibits certain advantages in knowledge extraction. When combined with emotional causal information, it generates knowledge that aligns more closely with the dialogue context.

\subsection{Effect of CoT Output Templates}
There are performance differences between different CoT output templates of the emotion and cause part in $R_2^{la}$. We explore various templates to analyze the accuracy of emotion recognition, the $macro\_F_1$ score of cause extraction, and the empathetic effect of the responses through manual scoring. As shown in Table~\ref{tab_5_template}, it demonstrates that analyzing emotions first yields better results than extracting causes first, as it aligns more closely with human reasoning. Additionally, template structures based on causal connectives such as "\textit{because}" achieved the highest emotional cause recognition performance and effectively enhanced the empathy of responses, which indicates that causal connectives can effectively uncover implicit causal relationships in dialogues  \citep{zhouhao}.

\section{Conclusion}
In this paper, we propose a novel cause-aware CoT fine-tuning  method for empathetic generation. Our proposed method leverages the designed CoT generation template to guide the model in conducting listener-aware cognitive inference, while also improving response effectiveness through fine-tuning. Additionally, we utilize emotional causes to further enhance the consistency between external knowledge and dialogue history. Detailed automatic and manual evaluation results demonstrate the state-of-the-art performance of our model.

\section{Limitations}
The limitations of our work can be summarized in the following two aspects. Firstly,we choose the strategy of manually crafting templates intuitively, while exploring additional prompt templates could potentially enhance empathetic effects. Secondly, empathetic responses also require the incorporation of more professional knowledge and skills in psychology.  Training LLMs with more empathetic dialogues and psychological counseling corpora could further advance the development of specialized empathetic conversational models.
% \bibliography{anthology,custom}
\bibliography{custom}

\appendix
\section{ Additional Case Study}
Table~\ref{tab_6_more_case} shows additional case study of generated
responses from SOTA baselines and our proposed methods.
\begin{table*}[t] % 't' 表示表格置于顶部
\centering % 居中显示表格
\begin{tabularx}{\textwidth}{@{}p{1.9cm}|X@{}} % 使用 tabularx 环境
\toprule
Emotion & Angry \\ 
\midrule
Context & Speaker: I'm so frustrated with my husband. He dismantled the bathroom to do repairs and still hasn't completed them six months later! \\
         & Listener: Six months!  I would be so mad!  What are you going to do about it? \\
         & Speaker: He promised he would complete it within a month and since it's been so long, I called a contractor today! \\
         & Ground-Truth: Good call!  I would have done the same thing! \\ \midrule
Knowledge
         & COMET: Intent to get something done; Need to have a job; Want to get a new one; Effect to gets yelled at by wife; feels frustrated. \\
         & Cause-oriented COMET: Intent to complete bathroom repairs; Need to bathroom repairs; Want to to finish repairs promptly; Effect: feels annoyed with husband; feels frustrated. \\ 
\midrule
ChatGPT & I understand your frustration. Bringing in a contractor seems like a practical solution. \\ \cline{2-2}
+Knowledge & That sounds tough. It's disappointing when projects take longer than expected. Bringing in a contractor seems like a practical move to get things back on track. \\ \cline{2-2}
+Cause-oriented Knowlegde & That's frustrating! \textbf{It's understandable to feel upset, especially when promises aren't kept.} Bringing in a contractor sounds like a good step toward resolving the situation efficiently. \\
\midrule
CFEG & SYS:A feels angry because he says "He dismantled the bathroom to do repairs and still hasn't completed them six months later!" I feel so sorry to hear that. Therefore, I will reassure him: Good idea! I can understand \textbf{how frustrating it must be to have the bathroom in disarray for so long. Bringing in a contractor seems like a practical solution to get things back on track.} \\
\bottomrule
\toprule
Emotion & Confident \\ 
\midrule
Context & Speaker:  I have a hard time having confidence in myself but when I got married and I had my hair and makeup done I felt like I was beautiful!  I had so much confidence in myself that day and it was great! \\
         & Listener: I bet you were beautiful on that day. :) I hope that confidence has carried on since then! \\
         & Speaker: I really try but it's hard to replicate that day for sure. \\
         & Ground-Truth: It would be hard to replicate that day for sure, but I bet you have a lot to be confident about. You got this! \\ \midrule
Knowledge
         & COMET: Intent to be happy; Need to have a good time; Want to have a good time; Effect to get married; feels confident. \\
         & Cause-oriented COMET: Intent to enhance self-confidence; Need to feel beautiful; Want to experience beauty; Effect: to get married; feels confident. \\ 
\midrule
ChatGPT & Your wedding day brought joy and confidence. Cherish that feeling! \\ \cline{2-2}
+Knowledge & You must be very happy. Getting married can make people more confident. \\ \cline{2-2}
+Cause-oriented Knowlegde & I can imagine your beauty and confidence. Congratulations on your marriage. Believe you can maintain it. \\
\midrule
CFEG & SYS:A feels confident because he says "when I got married and I had my hair and makeup done I felt like I was beautiful!  I had so much confidence" I feel so glad to hear that. Therefore, I will sympathize him: \textbf{Your wedding day sounds like a dream come true, filled with radiant confidence.} \textbf{Treasure those precious moments forever!} \\
\bottomrule
\end{tabularx}
\caption{Generated responses from baselines and our model. The general knowledge generated by COMET and ECPE oriented COMET is given in the Context row. \label{tab_6_more_case}}
\end{table*}
\end{document}